%% file: acl_paper.tex
\documentclass[11pt]{article}

\usepackage{acl}

\usepackage{times}
\usepackage{latexsym}

\usepackage[T1]{fontenc}

\usepackage[utf8]{inputenc}

\usepackage{microtype}

\usepackage{inconsolata}

\usepackage{graphicx}
\usepackage{subcaption}
\usepackage{booktabs} % for professional tables

\usepackage{amsmath}
\usepackage{amssymb}
\usepackage{mathtools}
\usepackage{amsthm}

\usepackage{array}
\usepackage{colortbl}    % For table coloring
\usepackage{multirow}
\usepackage{tabularx}

\theoremstyle{plain}

\theoremstyle{definition}

\theoremstyle{remark}

\usepackage[textsize=tiny]{todonotes}

\definecolor{thm}{RGB}{69, 53, 193}

\newcounter{assump}[section]

\usepackage[most]{tcolorbox}
\tcbuselibrary{skins, breakable}

\newtcolorbox{assumbox}[1][]{%
  colback=blue!5,       
  colframe=blue!50!black, 
  coltitle=white,       
  colbacktitle=blue!60!black, 
  boxrule=1.5pt,                 
  rounded corners,               
  fonttitle=\bfseries,  
  enhanced,
  breakable,
  attach boxed title to top left={yshift=-2mm,xshift=2mm},
  boxed title style={
    rounded corners,
    borderline west={0pt}{0pt}{white}, 
    borderline east={0pt}{0pt}{white},
    borderline north={0pt}{0pt}{white},
    borderline south={0pt}{0pt}{white},
  },
  before upper={\refstepcounter{assump}},
  #1
}

\title{Rethinking Memory as Continuously Evolving Connectivity}

\author{
  Jizhan Fang\textsuperscript{1,2},
  Buqiang Xu\textsuperscript{1},
  Zhixian Wang\textsuperscript{1},
  Haoliang Cao\textsuperscript{2},
  Xinle Deng\textsuperscript{1}, \\
  \textbf{Baohua Dong\textsuperscript{2}},
  \textbf{Hangcheng Zhu\textsuperscript{2}},
  \textbf{Ruohui Huang\textsuperscript{2}}, 
  \textbf{Gang Yu\textsuperscript{2}},
  \textbf{Ying Wei\textsuperscript{1}},\\
  \textbf{Guozhou Zheng\textsuperscript{1}},
  \textbf{Feiyu Xiong\textsuperscript{3}},
  \textbf{Haofen Wang\textsuperscript{4}},
  \textbf{Huajun Chen\textsuperscript{1}},
  \textbf{Ningyu Zhang\textsuperscript{1}\thanks{ \ \ Corresponding author.}}
  \\
  \textsuperscript{1}Zhejiang University
  \quad
  \textsuperscript{2}Alibaba Group 
  \quad
  \textsuperscript{3}MemTensor 
  \quad
  \textsuperscript{4}Tongji University\\
}

\begin{document}
\maketitle
\begin{abstract}
Existing memory-augmented LLM agents often treat memory as a static repository with pre-defined representations and fixed retrieval pipelines, which is brittle in dynamic agentic environments where feedback, task variation, and heterogeneous signals continuously reshape what should be remembered and how it should be connected. To address this, we propose \textbf{FluxMem}, a connectivity-evolving memory framework that models memory as a heterogeneous graph and progressively refines its topology through three stages: initial connection formation, feedback-driven refinement, and long-term consolidation. During execution, \textbf{FluxMem} repairs missing links, prunes interference, aligns abstraction granularity, and distills recurrent successful trajectories into reusable procedural circuits, guided by one metric for memory generalizability and evolutionary maturity. Across three fundamentally distinct benchmarks including \textbf{LoCoMo}, \textbf{Mind2Web}, and \textbf{GAIA}, \textbf{FluxMem} achieves consistent state-of-the-art performance, demonstrating strong adaptation and generalization in complex agentic environments. The code will be open-sourced in the near future\footnote{\url{https://github.com/zjunlp/LightMem}.}. 
\end{abstract}

\input{section/introduction}

\input{section/method_new}

\input{section/experiments}

\input{section/relatedwork}

\input{section/conclusion}

\input{section/limitations}

\input{section/Use_of_AI}
\bibliography{example_paper}

\appendix

\section{Experimental Details}
\label{appendix:experimental_details}

\subsection{Detailed Dataset Statistics}
\label{appendix:dataset_statistics}
\paragraph{LoCoMo.} The LoCoMo benchmark provides a specialized evaluation for long-context reasoning through $10$ extensive conversations, featuring an average of $588$ turns and $16,618$ tokens per dialogue. Our evaluation utilizes a total of $1,540$ human-annotated questions, covering $841$ single-hop, $282$ multi-hop, $321$ temporal reasoning, and $96$ open-domain questions. This distribution ensures a balanced assessment of both simple retrieval and complex logical synthesis.

\paragraph{Mind2Web.} This dataset serves as a large-scale testbed for generalist web agents, featuring $2,350$ open-ended tasks harvested from $137$ real-world websites across $31$ domains. The environmental complexity is significant, with an average page size of $1,135$ DOM elements and tasks requiring an average of $7.3$ discrete actions to complete. We specifically evaluate the model's generalization across three dimensions: cross-task, cross-website, and cross-domain scenarios.

\paragraph{GAIA.} Evaluation on the GAIA benchmark is conducted across $165$ curated tasks spanning three levels of increasing operational difficulty. The set includes $53$ Level-1 tasks focused on basic tool usage and retrieval, $86$ Level-2 tasks requiring multi-step planning and intermediate reasoning, and $26$ Level-3 tasks involving long-horizon execution and multi-modal integration.

\subsection{Implementation Details}
\label{appendix:implementation_details}
\paragraph{LoCoMo.}
For the LoCoMo benchmark, all retrieval-based methods utilize the \texttt{text-embedding-3-small} model for generating embeddings. The specific retrieval configurations for each baseline are as follows:
\textcircled{\scriptsize 1} \textbf{Zep} and \textcircled{\scriptsize 2} \textbf{Mem0} utilize their respective commercial APIs to retrieve memory contexts, extracting the top-10 relevant memories for each speaker, which are then integrated for response generation.
\textcircled{\scriptsize 3} \textbf{A-MEM} performs a global search to retrieve the top-40 overall entries.
\textcircled{\scriptsize 4} \textbf{MemoryOS} is implemented as a three-tier hierarchical system featuring exhaustive recall of all Short-Term Memory (STM) pages, a two-stage selection for Mid-Term Memory (MTM) (comprising the top-5 segments and top-10 dialogue pages), and the extraction of the top-10 relevant entries from Long-term Personal Memory (LPM).
\textcircled{\scriptsize 5} \textbf{Nemori} retrieves the top-10 episodic memories and top-20 semantic memories ($m=2k$).
\textcircled{\scriptsize 6} \textbf{LightMem} extracts the top-40 total entries through its retrieval pipeline.
\textcircled{\scriptsize 7} \textbf{MIRIX} utilizes a multi-agent active retrieval mechanism, generating automated topics to extract the top-10 entries from each of its six memory banks, totaling up to 60 entries.
\textcircled{\scriptsize 8} \textbf{EverMemOS} employs a hierarchical architecture based on memory scenes (MemScene) and memory cells (MemCell), utilizing hybrid retrieval (Dense+BM25) combined with a reranking strategy to extract the top-10 thematic scenes and top-10 narrative snippets.

\paragraph{Mind2Web.} 
For the Mind2Web benchmark, all baseline methods are evaluated under an offline protocol and follow their respective native evaluation pipelines,using GPT-4.1-mini and Gemini-2.5-flash as the base models.: \textcircled{\scriptsize 1} \textbf{AWM} is evaluated under two settings. In the original setting, we apply top-k = 5 candidate filtering on the current page. However, due to the large number of page elements, restricting the current page to top-k candidates discards substantial information and effectively reduces the action selection difficulty, which deviates from realistic web interaction. Therefore, in the realistic setting, we disable top-k filtering for the current page and instead provide the model with all candidate elements, including both positive and negative elements, to better simulate real-world conditions. In both settings, we keep the same configuration for visited-history elements, using top-k = 3 retrieval from previously visited pages.
\textcircled{\scriptsize 2} \textbf{ReasoningBank} The agent predicts actions with retrieved reasoning memories injected into the prompt, and scores are computed by the standard evaluation script over the test set.

\paragraph{GAIA.} 
For the GAIA benchmark, all methods are evaluated with a single run per task: \textcircled{\scriptsize 1} \textbf{Langfun} operates with Claude 3.7 Sonnet and associated models through Google's functional programming interface; \textcircled{\scriptsize 2} \textbf{OpenAI Deep Research} leverages o1/o3 series models with integrated chain-of-thought and reinforcement learning optimization; \textcircled{\scriptsize 3} \textbf{Magnetic-1} employs an o1-based open-source agentic infrastructure; \textcircled{\scriptsize 4} \textbf{Agent KB} incorporates an offline knowledge base pre-indexed with domain expertise; \textcircled{\scriptsize 5} \textbf{Alita} uses multi-model orchestration (Claude 3.7 Sonnet + GPT-4o) with dynamic model selection; \textcircled{\scriptsize 6} \textbf{Smolagents} implements a lightweight two-agent architecture (manager + tool agent). Our evaluations of \textcircled{\scriptsize 7} \textbf{Flash-Searcher} (baseline), \textcircled{\scriptsize 8} \textbf{MemEvolve}, and \textcircled{\scriptsize 9} \textbf{FluxMem} are conducted within the Flash-Searcher framework, configured with 40 maximum steps, 8-step planning intervals. Flash-Searcher operates without memory augmentation (32,768 completion tokens per call). MemEvolve employs meta-evolved architectures through 3 iterations, 60 trajectories per round with 40 new + 20 reused tasks, (K=1) survivor, (S=3) descendants. 

\end{document}

%% file: section/introduction.tex
\section{Introduction}

\label{sec:intro}

For long-horizon agents, memory mechanism~\citep{zhang2025survey, hu2026memoryageaiagents} plays a central role~\citep{mei2025surveycontextengineeringlarge}, by distilling useful factual information, reusable experiences and skills from the agent's past interaction trajectories~\citep{packer2024memgptllmsoperatingsystems, wei2025evomemorybenchmarkingllmagent, zhang2026memskilllearningevolvingmemory}, storing them in diverse memory forms, and retrieving relevant memories when similar tasks arise to support downstream problem solving and agent evolving~\citep{qi2026individualintelligencesurveyingcollaboration, gao2026surveyselfevolvingagentswhat, qiu2025alitageneralistagentenabling, wang2026skillxautomaticallyconstructingskill,ye2026autodreamerlearningofflinememory}. 
For long-horizon agents, memory effectiveness ultimately depends on whether the \emph{most useful memories} can be accessed at each decision step, as sufficiently useful memory context substantially improves subtask success. 

\begin{figure}[!t]
    \centering
    \includegraphics[width=\linewidth]{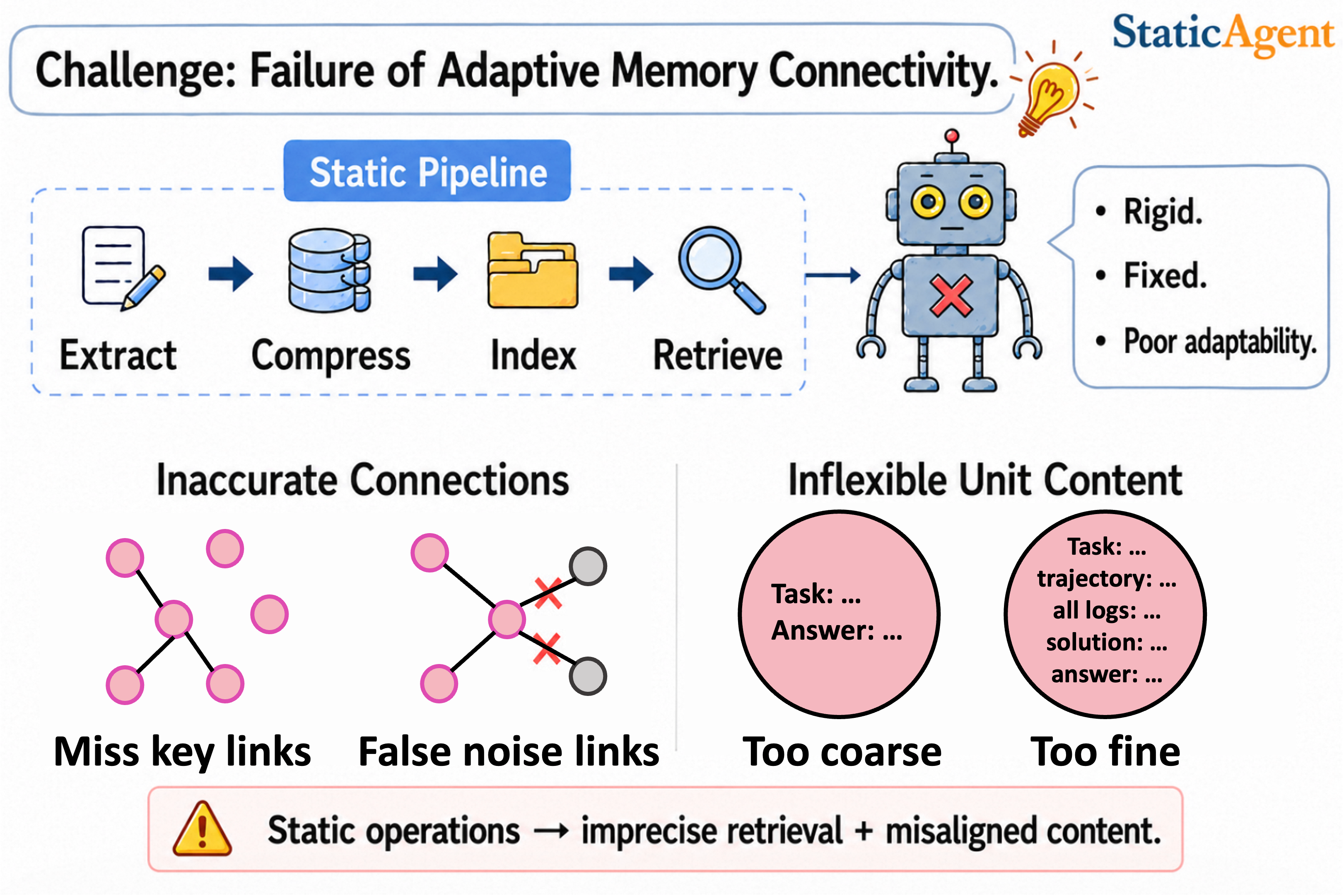}
    \caption{The failures of static memory systems.}
    \vspace{-5mm}
    \label{fig:intro}
\end{figure}

We formalize such usefulness as a problem of \emph{memory connectivity}. 
Drawing from cognitive science~\citep{hebb2005organization, frankland2005organization}, we define memory as the long-term sedimentation of \textbf{memory units} and their \textbf{connections}, continuously shaped through environmental interaction. 
Mirroring human cognitive processes, this structural evolution operates on two levels. 
\emph{At the unit level}, the brain generates new units for novel information and continuously reshapes existing units by modifying their internal content. 
This ensures that each memory unit dynamically integrates new experiences and refines its semantic representation. 
\emph{At the connection level}, operations are strictly task-centric, the system establishes links between co-activated units to form functional associations, and prunes links that prove irrelevant, maintaining an efficient associative network. 
Through repeated task execution and environmental feedback, these localized updates gradually consolidate into stable, large-scale regions of interconnected nodes and edges. 
Rather than static storage, memory thus emerges as a self-organizing structure that its memory units and connections continuously adapt and evolve over time~\citep{kelly2005human}. 

\paragraph{Challenges.} 
\textbf{First, \emph{Failure of Adaptive Memory Connectivity}.} 
Existing methods predominantly rely on static, hand-crafted pipelines~\citep{yang2026plugmemtaskagnosticpluginmemory, chhikara2025mem0buildingproductionreadyai, fang2025lightmem, suzgun-etal-2026-dynamic}. 
By hardcoding memory operations, they assume rigid designs and fixed operations generalize across tasks. 
However, such static paradigms cannot establish optimal memory structures for diverse scenarios or dynamically refine them based on environmental feedback~\citep{zhang2025memevolvemetaevolutionagentmemory, chen2026memprivacyprivacypreservingpersonalizedmemory}. This inflexibility creates bottlenecks at both the connection and unit levels:
\textbf{(1) \emph{Inaccurate Memory Connections}.} 
This inaccuracy primarily manifests during \textit{memory retrieval}. 
It leads to two concrete failures: \textit{under-connection}, where critical links are missed due to retrieval imprecision, depriving the agent of essential context, and \textit{over-connection}, where irrelevant associations are indiscriminately retrieved, introducing noise and hallucinations~\citep{jiang2025personamemv2personalizedintelligencelearning, chen2026halumemevaluatinghallucinationsmemory}. Fundamentally, static pipelines lack the dynamic adaptability required for precise connection formation and access. 
\textbf{(2) \emph{Inflexible Memory Unit Content}.} 
Existing systems represent memory units at a single, predefined level of abstraction. 
When unit content is misaligned, either excessively coarse, losing critical execution details, or overly fine, obscuring high-level structural patterns, the memory unit fails to adaptively integrate new experiences. 

\textbf{Second, \emph{Failure of Memory Connection Consolidation}.} 
While existing systems preserve task trajectories~\citep{fang2026mempexploringagentprocedural, ouyang2025reasoningbankscalingagentselfevolving, tang2025chemagentselfupdatinglibrarylarge}, they treat memories as isolated instances rather than progressively consolidating them. True consolidation requires localized updates to coalesce through feedback into stable, large-scale associative regions. Lacking this mechanism, agents repeatedly reconstruct similar associations instead of internalizing enduring structural patterns, preventing memory networks from self-organizing into optimal configurations.

% \textbf{Inspiration.} 
% From theories of memory consolidation in human neuroscience, memory can be viewed as the evolution of neural connections through a multi-timescale process. 
% \textbf{(1) Initial Memory Encoding.} 
% When humans first learn a new task, the brain forms sparse and fragile neural connections linking newly acquired information with existing memories of related concepts, experiences or skills. These tentative associations are easily disrupted, often leading to unstable and error-prone execution~\citep{hebb2005organization}. 
% \textbf{(2) Dynamic Refinement via Feedback.} 
% To gradually form robust memory structures, the brain continuously refines these connections through internal predictions and external feedback~\citep{schultz1997neural}. 
% Specifically, successful associations are reinforced while misleading ones fade away, and new links are formed or eliminated as needed, progressively reshaping the connectivity into stable and efficient connections~\citep{kandel2001molecular}. 
% \textbf{(3) Long-term Consolidation.} 
% Over longer timescales, repeated reinforcement consolidates these pathways into robust neural circuits. 
% For example, after repeatedly practicing a familiar task, humans no longer need to consciously reconstruct every intermediate reasoning step. 
% As a result, well-learned tasks can often be executed rapidly and almost automatically with minimal cognitive effort, reflecting the transition from fragile exploratory connections to stable and efficient neural structures~\citep{frankland2005organization, kelly2005human}. 

\paragraph{Method.} 
To address these challenges, we propose \textbf{FluxMem}, a connectivity-evolving framework that models memory as a dynamically editable heterogeneous graph across semantic, episodic, and procedural layers. 
Context is formalized as an activated subgraph refined through a three-stage evolutionary pipeline. 
\emph{(1) Initial Connection Formation} rapidly establishes tentative cross-layer associations for novel tasks. 
\emph{(2) Feedback-Driven Refinement} employs a closed-loop mechanism to iteratively edit subgraph topology, creating missing links, pruning interference, or conditionally bypassing memory until execution succeeds. 
\emph{(3) Long-Term Consolidation} clusters successful trajectories to induce stable procedural circuits, monitored by a convergence maturity metric. 
As high-utility pathways crystallize, recurring tasks bypass redundant retrieval and directly activate mature subgraphs. 
This pipeline transforms static memory storage into a self-optimizing connectivity substrate that continuously adapts to evolving task demands.

\paragraph{Results.} 
We evaluate \textbf{FluxMem} on three benchmarks covering distinct task scenarios to evaluate generalization: LoCoMo~\citep{maharana2024evaluating} (long-context reasoning), Mind2Web~\citep{deng2023mind2web} (real-world web navigation), and GAIA~\citep{mialon2023gaia} (general assistant tasks). \textbf{FluxMem} achieves state-of-the-art performance across all three benchmarks.
On LoCoMo, \textbf{FluxMem} reaches 95.06 average accuracy, above the Full Context baseline (81.23). On Mind2Web in the realistic setting (no manual element filtering), \textbf{FluxMem} improves Cross-Task success rate to 8.1, more than AWM~\citep{wang2024agent} (3.6). 
On GAIA, \textbf{FluxMem} increases the average success rate from 52.12 to 64.85 on \textit{Kimi K2} (+12.73\% absolute) compared with the \textit{Flash-Searcher}~\citep{qin2025flashsearcherfasteffectiveweb} baseline, and also surpasses the strong \textit{MemEvolve}~\citep{zhang2025memevolvemetaevolutionagentmemory} baseline.
% Ablations further show that reflection (Stage II) is the main driver for reasoning-heavy LoCoMo (removing it drops performance by nearly 10 points), while maturity-based evolution (Stage III) contributes most for long-horizon web navigation (removing it roughly halves success rate in representative settings).
% Our contributions are as follows:
% \begin{itemize}
%     \item We propose \textbf{FluxMem}, an evolutionary memory framework that models agent memory as a four-layer heterogeneous graph (Source--Evidence--Episode--Anchor) and continuously restructures it through interaction and feedback.
%     \item We introduce a three-phase mechanism, task space exploration with cluster expansion and anchor inheritance, failure-driven reflection for adaptive evidence/anchor selection, and offline evolutionary, to continuously transform noisy trajectories into task-solving anchors.
%     \item We develop two metrics, Task-Evidence Distance (TED) and Anchor Evolution Maturity Score (AEMS), to guide task-level generalization and to quantify when memory evolution should terminate; extensive experiments on LoCoMo, Mind2Web and GAIA demonstrate sota performance and validate the role of each stage.
% \end{itemize}

%% file: section/method_new.tex
\section{FluxMem Memory Architecture}

\begin{figure*}[t]
    \centering
    \includegraphics[width=1\textwidth]{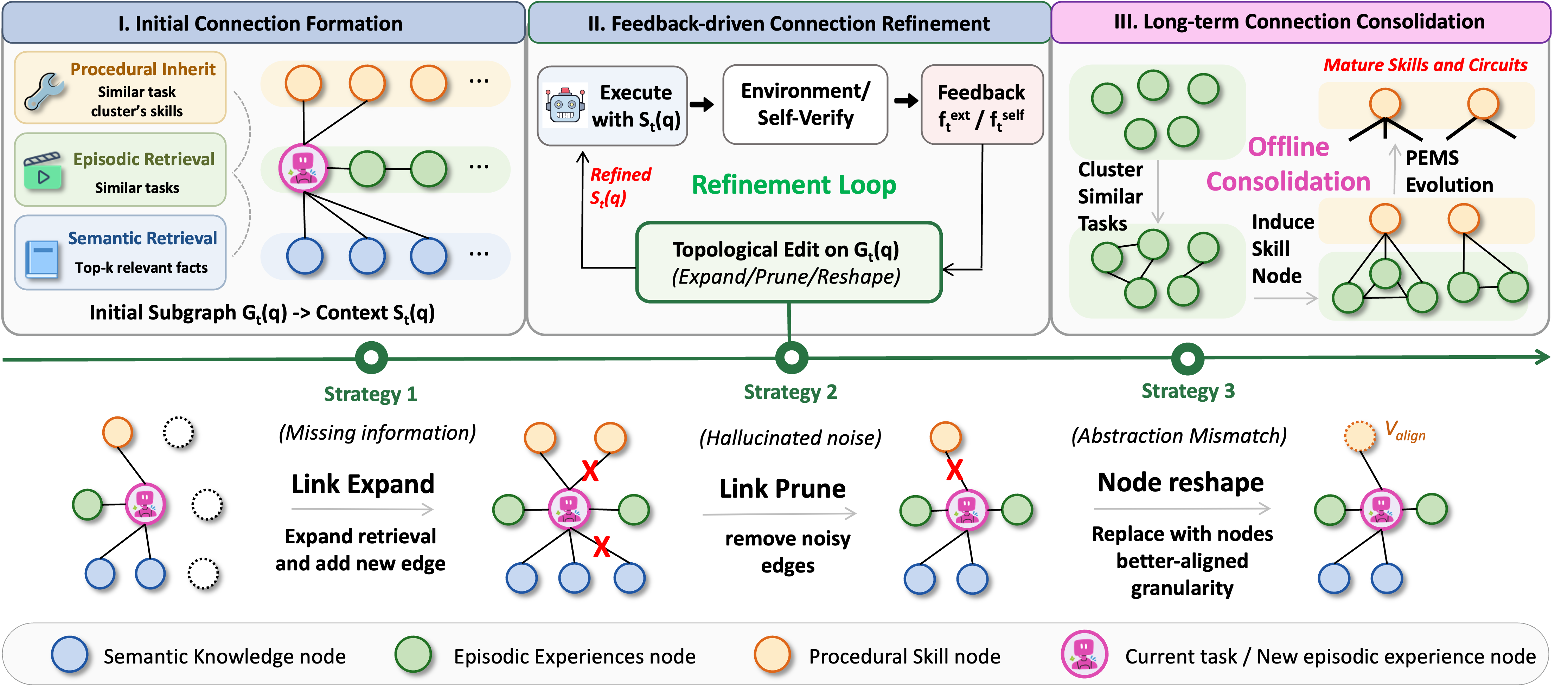}
    \vspace{-0.4cm}
    \caption{The \textbf{FluxMem} architecture. Stages I and II operate online at a step-wise granularity. Stage III is conducted offline, aiming for immediate performance optimization and long-term memory consolidation, respectively. }
    \label{fig:architecture}
\end{figure*}

\subsection{Three-Layer Memory Graph}
We model \textbf{FluxMem} as a heterogeneous graph $\mathcal{G}=(\mathcal{V}, \mathcal{E})$, where the node set $\mathcal{V}$ comprises three functional layers. \textcircled{\scriptsize1} \textbf{Semantic Knowledge} $\mathcal{V}_{\text{sem}}$ stores static factual knowledge that provides evidential support (e.g., knowledge documents or their corresponding chunks). 
\textcircled{\scriptsize2} \textbf{Episodic Experiences} $\mathcal{V}_{\text{epi}}$ records concrete state-action trajectories (e.g., debugging logs or tool-use sequences). 
\textcircled{\scriptsize3} \textbf{Procedural Skills} $\mathcal{V}_{\text{proc}}$ encapsulates distilled reasoning templates (e.g., multi-step planning heuristics). 

Among the 3 layers, $\mathcal{V}_{\text{epi}}$ serves as the operational nexus that orchestrates the interplay between static knowledge and distilled skills. 
Each node $v_{\text{epi}}^{(q)} \in \mathcal{V}_{\text{epi}}$ represents a specific task $q$ and records its full step-by-step trajectory $\tau_q = \{(o_t, a_t)\}_{t=1}^{T}$. 
The three layers are linked in a bottom-up order through two types of edges in $\mathcal{E}$. 
First, during task execution, the agent retrieves relevant facts from $\mathcal{V}_{\text{sem}}$ to explain its current observation and decide the next step. 
This creates the edge set $\mathcal{E}_{\text{ground}} \subseteq \mathcal{V}_{\text{sem}} \times \mathcal{V}_{\text{epi}}$, where an edge $(v_{\text{sem}}, v_{\text{epi}}^{(q)})$ simply indicates that a specific fact provides evidential support for a step in task $q$. 
Second, after completing one or more similar tasks, the agent identifies common patterns in these trajectories and summarizes them into reusable skills. 
This forms the edge set $\mathcal{E}_{\text{distill}} \subseteq \mathcal{V}_{\text{epi}} \times \mathcal{V}_{\text{proc}}$, where an edge $(v_{\text{epi}}^{(q)}, v_{\text{proc}})$ shows that a skill is distilled from past experiences. 
Once created, $v_{\text{proc}} \in \mathcal{V}_{\text{proc}}$ can be used to guide the agent in future tasks. 

In \textbf{FluxMem}, the semantic layer is sourced directly from the environment, encompassing raw inputs such as dialogue histories and tool API documentation. Episode nodes are instantiated individually for each task, while procedural skill nodes are subsequently induced in section~\ref{longterm}.

% 仔细介绍三个层的schema定义以及图的形态

\subsection{Context as Dynamically Induced Connectivity}

At each step $t$ of task $q$, \textbf{FluxMem} constructs the agent's context $S_t(q)$. 
The system dynamically selects a task-specific subset of nodes to form a local subgraph $\mathcal{G}_t(q) = (\mathcal{V}_t, \mathcal{E}_t) \subset \mathcal{G}$, where $\mathcal{V}_t = \mathcal{V}_t^{\text{sem}} \cup \mathcal{V}_t^{\text{epi}} \cup \mathcal{V}_t^{\text{proc}}$ contains the activated memory nodes from the three layers. 
\[
S_t(q) = \text{Concat}( q, \; \text{Obs}_t, \; \mathcal{V}_t^{\text{sem}}, \; \mathcal{V}_t^{\text{epi}}, \; \mathcal{V}_t^{\text{proc}} ),
\]
In this formula, $\text{Obs}_t$ describes the observations. 
Under this formulation, optimizing the working context is equivalent to performing targeted topological edits on $\mathcal{G}_t(q)$, as the prompt content is strictly determined by the activated node set and edge connections. 
Consequently, the adaptation pipeline systematically evolves $\mathcal{G}_t(q)$ from fragile tentative links into robust, task-optimized circuits through three sequential stages.

\section{Three-Stage Memory Evolution}

\textbf{FluxMem} comprises three stages. Stages I and II operate online at a step-wise granularity during task execution. At each time step $t$, Stage I is executed first to generate $S_t(q)$, which is immediately processed by Stage II to yield the refined context $S_t'(q)$. Stage III is conducted offline.

\subsection{Stage I: Initial Connection Formation}

\paragraph{Semantic Connection Retrieval.} 
At time step $t$, given the current observation $o_t$, the system establishes initial associations between $o_t$ and supporting factual knowledge by querying the semantic layer $\mathcal{V}_{\text{sem}}$. 
We compute a hybrid relevance score for each candidate $v \in \mathcal{V}_{\text{sem}}$ by fusing dense embedding similarity, sparse lexical matching, and LLM-based verification:
\[
\text{Score}(v, o_t)
=
\frac{\mathbf{v} \cdot \mathbf{o}_t}
{\|\mathbf{v}\| \|\mathbf{o}_t\|}
+ \text{BM25}(v, o_t)
\]
\[
+ \text{LLM}_{\text{ver}}(v, o_t).
\]
The top-$k$ nodes instantiate $\mathcal{V}_{t}^{\text{sem}}$, with directed edges $\mathcal{E}_{t}^{\text{sem}} = \{(v_{t}, v) \mid v \in \mathcal{V}_{t}^{\text{sem}}\}$ established to link them to the current step anchor $v_t$.

\paragraph{Episodic Connection Retrieval.} 
To draw on experience from past similar tasks, we query the episodic layer $\mathcal{V}_{\text{epi}}$ for the $k$ most relevant past episodes using embedding similarity. 
\[
\mathcal{V}_{t}^{\text{epi}} = \text{TopK}_{u \in \mathcal{V}_{\text{epi}}} \big\{ \cos(\mathbf{u}, \mathbf{o}_t) \big\}.
\]

\paragraph{Procedural Connection Inheritance.} 
Based on the retrieved episodes $ \mathcal{V}_{t}^{\text{epi}} $, we collect applicable skills by traversing existing distillation connections. Specifically, we select all skill nodes $v_{\text{proc}}$ that are linked to any retrieved episode via $\mathcal{E}_{\text{distill}}$:
$$
\mathcal{V}_{t}^{\text{proc}} = \bigcup_{v_{\text{epi}} \in \mathcal{V}_{t}^{\text{epi}}} \big\{ v_{\text{proc}} \mid (v_{\text{epi}}, v_{\text{proc}}) \in \mathcal{E}_{\text{distill}} \big\}.
$$
The retrieved facts, episodes, and skills together form the initial step-local subgraph $\mathcal{G}_t = (\mathcal{V}_t, \mathcal{E}_t)$, where $\mathcal{V}_t = \mathcal{V}_{t}^{\text{sem}} \cup \mathcal{V}_{t}^{\text{epi}} \cup \mathcal{V}_{t}^{\text{proc}}$. This subgraph is serialized into the initial step context $S_t$. Although this provides a complete starting point for the current step's reasoning, the selected connections are preliminary and will be refined in Stage II.

\subsection{Stage II: Feedback-Driven Connectivity Refinement}
Following the initial retrieval, the system addresses structural misalignments through a feedback-driven refinement loop. At step $t$, upon receiving execution feedback $f_t$ (from environmental signals or self-verification), the agent attributes reasoning failures to either \emph{connection-level} or \emph{unit-level} flaws and applies targeted edits to $\mathcal{G}_t$.

\paragraph{Connection-Level Refinement.} To resolve inaccurate memory connections, the system dynamically adjusts the associative topology based on feedback attribution. 
(i) \textbf{Link Expansion for Under-Connection.} If $f_t$ indicates missing critical context, the system identifies semantically proximate but unactivated nodes $v_{\text{new}} \in \mathcal{V} \setminus \mathcal{V}_t$ and establishes new task-centric edges via $\mathcal{E}_t \leftarrow \mathcal{E}_t \cup \{(v_t, v_{\text{new}})\}$. 
(ii) \textbf{Link Pruning for Over-Connection.} If $f_t$ reveals context congestion or hallucinated guidance, the system identifies distractor edges $\mathcal{E}_{\text{noise}} \subset \mathcal{E}_t$ and severs them via $\mathcal{E}_t \leftarrow \mathcal{E}_t \setminus \mathcal{E}_{\text{noise}}$, isolating $v_t$ from irrelevant associations.

\paragraph{Unit-Level Refinement.} To overcome inflexible memory unit content, the system dynamically reshapes internal representations when granularity misalignment impedes step-level reasoning. 
(iii) \textbf{Content Reshaping for Granularity Alignment.} When retrieval is sufficient but the unit abstraction mismatches current demands (e.g., overly coarse for precise execution or overly fine for high-level planning), the system adaptively modifies the internal content of $v_{\text{old}} \in \mathcal{V}_t$. This involves either expanding $v_{\text{old}}$ with finer-grained execution details or abstracting redundant components to elevate its semantic level, yielding a refined unit $v_{\text{align}}$. The local subgraph is updated by replacing $v_{\text{old}}$ with $v_{\text{align}}$ while preserving established connections.

After applying the targeted edits, the refined subgraph $\mathcal{G}_t' = (\mathcal{V}_t', \mathcal{E}_t')$ is serialized into the updated context $S_t'$ for subsequent reasoning. The loop terminates upon execution success or reaching a predefined refinement rounds $T$.

\subsection{Stage III: Long-Term Connection Consolidation}
\label{longterm}

\paragraph{Episodic Clustering and Skill Induction.} 
Upon task completion, trajectories are committed as episodic nodes $v_{\tau} \in \mathcal{V}_{\text{epi}}$. During offline consolidation, the system first partitions $\mathcal{V}_{\text{epi}}$ into $M$ clusters $\{\mathcal{C}_m\}_{m=1}^M$ based on semantic trajectory similarity, computed via cosine distance between episode embeddings $\mathbf{u}_{\tau}$. For each cluster $\mathcal{C}_m$, an LLM-based induction operator extracts the skills or reasoning pattern shared across episodes, abstracting them into a new procedural skill node $v_{\text{proc}}^{(m)} \in \mathcal{V}_{\text{proc}}$. 

\paragraph{PEMS-Guided Iterative Consolidation.} 
Since previous initial skill induction is one-way and may produce invalid skills, we verify and optimize them through a closed-loop refinement process guided by iterative evolution. At each iteration $k$, the system re-runs the source episodes $\mathcal{C}_m$ that generated each skill $v_{\text{proc}}$, using the current skill version as guidance. We then compute the Procedure Evolution Maturity Score (PEMS) for every skill:
\[
\text{PEMS}^{(k)} = \frac{\eta\big(\mathcal{V}_{\text{proc}}^{(k)}\big)}{\log \ell\big(\mathcal{V}_{\text{proc}}^{(k)}\big)} \times \Big(1 - \delta\big(\mathcal{G}_{\text{cons}}^{(k)}, \mathcal{G}_{\text{cons}}^{(k-1)}\big)\Big),
\]
where $\eta^{(k)}$ is the average success rate of the source episodes under the current skill, $\ell^{(k)}$ is the token length of the skill text, and $\delta^{(k)}$ measures the embedding difference between the current and previous skill versions. Based on the execution results, the LLM directly rewrites low-scoring skills to fix logical errors or remove redundant content. This test-score-refine cycle repeats until the score improvement $\Delta \text{PEMS}^{(k)}$ falls below $\epsilon$. At that point, the skills are validated as both highly useful and concise, and the offline consolidation ends.

%% file: section/experiments.tex
\input{table/main_results_locomo}

\section{Experiments}

% 分区实验(整个图建完之后根据簇间距离指标可视化来证明聚类的效果)
% EDM随着轮数的变化情况，以及收敛的特性
% 消融实验，三个步骤，效果的变化

\subsection{Experimental Setup}

\paragraph{Datasets \& Baselines.}
We evaluate the proposed framework across three challenging benchmarks. 
\textbf{LoCoMo}~\citep{maharana2024evaluating} provides a comprehensive evaluation for long-context reasoning, we compare \textbf{FluxMem} against several representative baselines of conversational memory modeling: 
\textcircled{\scriptsize 1} Zep~\citep{rasmussen2025zeptemporalknowledgegraph},
\textcircled{\scriptsize 2} Mem0~\citep{chhikara2025mem0buildingproductionreadyai},
\textcircled{\scriptsize 3} A-Mem~\citep{xu2025mem},
\textcircled{\scriptsize 4} MemoryOS~\citep{kang2025memory},
\textcircled{\scriptsize 5} Nemori~\citep{nan2025nemori},
\textcircled{\scriptsize 6} LightMem~\citep{fang2025lightmem},
\textcircled{\scriptsize 7} MIRIX~\citep{wang2025mirix}, and
\textcircled{\scriptsize 8} EverMemOS~\citep{hu2026evermemos}.
\textbf{Mind2Web}~\citep{deng2023mind2web} serves as a testbed for web navigation, we compare \textbf{FluxMem} against representative baselines: \textcircled{\scriptsize 1} AWM~\citep{wang2024agent} and \textcircled{\scriptsize 2} Reasoning Bank~\citep{ouyang2025reasoningbankscalingagentselfevolving}. 
For general assistant tasks, we employ \textbf{GAIA}~\citep{mialon2023gaia} to benchmark against a wide array of frameworks:
\textcircled{\scriptsize 1} OpenAI Deep Research~\citep{deepresearch},
\textcircled{\scriptsize 2} Langfun~\citep{Peng_Langfun_2023},
\textcircled{\scriptsize 3} Magnetic-1~\citep{fourney2024magenticonegeneralistmultiagentsolving},
\textcircled{\scriptsize 4} Agent KB~\citep{tang2025agentkbleveragingcrossdomain},
\textcircled{\scriptsize 5} smolagents~\citep{smolagents},
\textcircled{\scriptsize 6} Alita~\citep{qiu2025alitageneralistagentenabling},
\textcircled{\scriptsize 7} Flash-Searcher~\citep{qin2025flashsearcherfasteffectiveweb}, and
\textcircled{\scriptsize 8} MemEvolve~\citep{zhang2025memevolvemetaevolutionagentmemory}. 

\paragraph{Metrics.}
For \textbf{LoCoMo}, we report the LLM-as-a-judge (LMJ) score. 
For \textbf{Mind2Web}, we evaluate action-level accuracy with Element Accuracy (EA), Action F1 (AF1), Step Success Rate (SSR), and report overall Success Rate (SR) for completing a full navigation task. 
For \textbf{GAIA}, we use Success Rate across Level 1--3 to measure end-to-end task completion under increasing difficulty. 
Further statistics and experimental details about baselines are provided in Appendix~\ref{appendix:dataset_statistics}.

% \paragraph{Experimental Details.}
% Across all three datasets, the key parameters for \textbf{FluxMem} are set as follows: 
% in Stage I, we perform $n=5$ rewrites or expansions for each query; 
% in Stage II (Error Correction and refinement), we set $T_{\max}=1$ for LoCoMo and $T_{\max}=5$ for Mind2Web and GAIA; 
% in Stage III, the threshold $\tau$ is set to $10$, and the sensitivity threshold $\epsilon$ for maturity $\Delta \mathcal{M}$ is set to $0.01$.

\input{table/main_results_mind2web}
\input{table/main_results_gaia}

\subsection{Main Results}
\label{sec:main_results}

\paragraph{Superiority in Long-Context Reasoning.}
As shown in Table~\ref{tab:locomo_results}, \textbf{FluxMem} sets a new state-of-the-art across all sub-categories on the LoCoMo benchmark. 
With the \textit{GPT-4.1-mini} backbone, FluxMem achieves an outstanding average LMJ score of \textbf{95.06}, significantly surpassing \textit{Full Context} (\textbf{81.23}) and the strongest specialized memory system \textit{EverMemOS} (\textbf{93.05}). 
This performance gap is even more pronounced when using \textit{Qwen3-30B-A3B-2507-Instruct} backbone, where \textbf{FluxMem} maintains a high average LMJ of \textbf{93.44}, while the next best baseline \textit{Full Context} drops to \textbf{74.87}. 
% These results suggest that our evolutionary graph distillation not only filters noise more effectively than raw context but also exhibits superior robustness across different model architectures.

\paragraph{Robust Performance in Web Navigation.}
As shown in Table~\ref{tab:mind2web_results}, the evaluation on Mind2Web highlights the adaptability of \textbf{FluxMem} in noisy, real-world web environments. 
In the realistic setting without manual element filtering ($‡\ddagger
‡$), our framework demonstrates consistent improvements across both backbone models. 
With \textit{GPT-4.1-mini}, \textbf{FluxMem} achieves a Success Rate (SR) of \textbf{8.1} in \textit{Cross-Task} scenarios, more than \textit{AWM} baseline (\textbf{3.6}). 
This trend is further reinforced with \textit{Gemini-2.5-flash}, where \textbf{FluxMem} reaches an even higher SR of \textbf{9.6} in \textit{Cross-Task} evaluation, substantially outperforming \textit{AWM} (\textbf{5.6}). Across all sub-categories and model backbones, \textbf{FluxMem} consistently yields highest SSR and AF1 scores. 
% This demonstrates that the crystallized memory anchors provide precise functional guidance that remains stable even when the agent encounters unfamiliar web structures.

\paragraph{State-of-the-Art on Generalist Assistant Tasks.}
On GAIA benchmark, \textbf{FluxMem} demonstrates exceptional gains over the high-performance \textit{Flash-Searcher} baseline and the meta-evolutionary system \textit{MemEvolve} in Table~\ref{tab:gaia_results}. When utilizing \textit{Kimi K2}, our framework boosts the average success rate from \textbf{52.12} to \textbf{64.85}, achieving a remarkable absolute improvement of \textbf{12.73\%}. In high-complexity tasks (\textit{Level 3}), \textbf{FluxMem} reaches a success rate of \textbf{53.85} with \textit{GPT-5-mini}, effectively matching or exceeding the capabilities of much larger closed-source agent frameworks. 

% The consistent performance of \textbf{FluxMem} across various tasks and backbone models demonstrates its capability to support diverse agent scenarios. 
% Furthermore, its mechanism of leveraging environmental feedback to guide evolution effectively ensures robust agent performance. 

\begin{figure*}[t]
    \centering
    \includegraphics[width=1\textwidth]{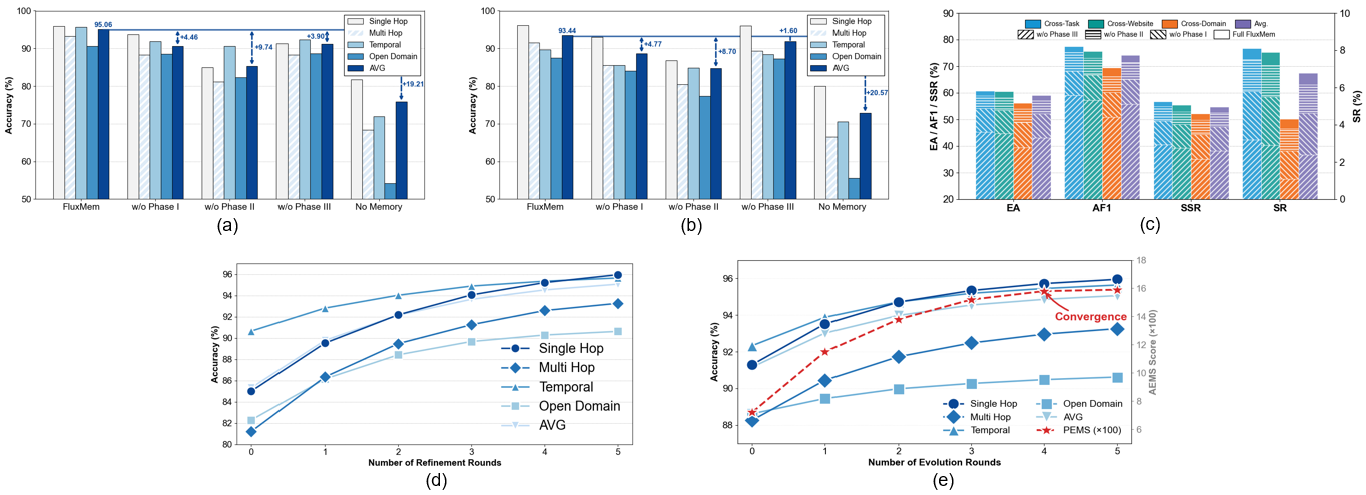}
    \vspace{-0.4cm}
    \caption{Detailed analysis of FluxMem components and evolution dynamics: 
    (a) Ablation study of different stages on LoCoMo using GPT-4.1-mini. 
    (b) Ablation study of different stages on LoCoMo using Qwen3-30B-A3B-2507-Instrcut. 
    (c) Ablation study of different stages on Mind2Web using GPT-4.1-mini. 
    %(d) Impact of query rewrite counts and expansion strategies on golden evidence retrieval success rate. 
    (d) Performance improvement across sub-tasks relative to the number of refinement rounds in Stage II. 
    (e) Model accuracy trends and convergence of the Procedure Evolution Maturity Score (PEMS) across evolution rounds.}
    \label{fig:analysis}
\end{figure*}

\subsection{Ablation Study}

We conduct ablation studies on LoCoMo and Mind2Web to evaluate the contribution of three stages, as shown in Figure~\ref{fig:analysis} (a),(b) and (c). 

On LoCoMo dataset, Stage II (Feedback-Driven Refinement) proves to be the most critical component. 
For GPT-4.1 mini, removing Stage II leads to a substantial decrease in the average LMJ score, dropping from 95.06\% to 85.32\%. 
A similar trend is observed with Qwen3-30B-A3B, where the average score falls from 93.44\% to 84.74\% upon the exclusion of Stage II, while other two ablations show relatively smaller impacts. 
% These results suggest that for reasoning-intensive tasks like LoCoMo, the immediate feedback and self-correction mechanism in Stage II are essential for maintaining agent performance. These results suggest that for reasoning-intensive tasks like LoCoMo, the immediate feedback and self-correction mechanism in Stage II are essential for maintaining agent performance.
In memory-centric scenarios like LoCoMo, where all required evidence can be directly retrieved or simply inferred from the provided context, tasks rely more on accurate recall than complex reasoning. 
Consequently, the Stage II refinement mechanism, which expands retrieval or prunes irrelevant memory nodes, yields substantial performance gains by helping the agent find the correct facts. 
In such settings, refining the semantic knowledge layer proves highly effective, while the procedural skill layer contributes relatively less. 

In contrast, on the Mind2Web dataset, Stage III (Long-Term Consolidation) emerges as the primary performance driver. 
For instance, removing Stage III for GPT-4.1-mini causes a drastic performance drop (e.g., the success rate on the first sub-category falls from 8.1\% to 3.2\%), while the impact of removing Stage II is relatively moderate. 
This disparity suggests that for complex, multi-step web navigation tasks requiring strong reasoning, the extraction of skills and evolution of skill nodes in Stage III are more vital than short-term refinement.

\subsection{Analysis of Iterative Refinement}
We analyze the impact of the number of refinement iterations in Stage II on performance.
The number of refinement rounds ($T$) in Stage II serves as a critical scaling factor. 
We evaluate this effect on the LoCoMo by varying $T$ from 0 to 5, as shown in Figure~\ref{fig:analysis}(d). 
The results demonstrate a consistent and monotonic improvement across all sub-categories and the overall score. 
Without refinement ($T=0$), the agent achieves an average score of 85.32\%. 
By $T=5$, the average performance reaches 95.06\%. 
This steady gain suggests that the refinement mechanism allows the agent to refine connections and find more useful factual evidences. 
The diminishing returns observed between $T=4$ and $T=5$ (an improvement of only 0.54\%) indicate that the agent's performance begins to saturate as it approaches the optimal evidence path.

\begin{figure*}[t]
    \centering
    \includegraphics[width=1\textwidth]{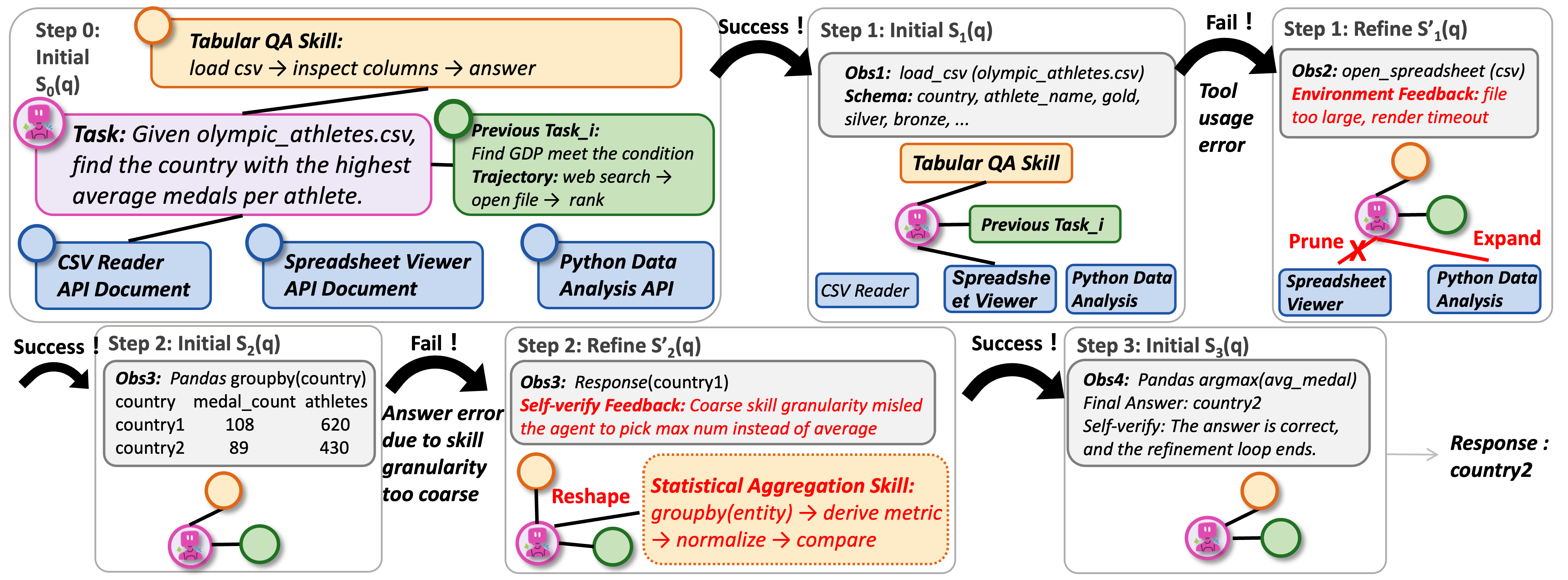}
    \vspace{-0.4cm}
    \caption{Case Study. The key points have been highlighted in red.}
    \label{fig:case}
\end{figure*}

\subsection{Analysis of Memory Evolution and Convergence}
At this point, the sensitivity threshold $\epsilon$ can be utilized to terminate the evolution process.

As shown in Figure~\ref{fig:analysis}(e), on the LoCoMo dataset, while Stage III provides a performance boost (from 91.16\% at round 0 to 95.06\% at round 5), the gains are more moderate compared to Stage II. This aligns with our observation that for fact-oriented tasks, the primary role of Stage III is to summarize and stabilize background knowledge. More importantly, we observe a clear convergence trend in the PEMS metric (scaled by a factor for visibility). The PEMS increases from 0.072 to 0.158 within the first four rounds and stabilizes at 0.159 by round 5. This convergence indicates that the memory maturity mechanism effectively identifies when the anchor nodes have reached a stable state of knowledge representation. At this point, the sensitivity threshold $\epsilon$ can be utilized to terminate the evolution process, thereby preventing redundant computations once the memory has captured the essential task-related insights. 

\subsection{Case Study}
Figure~\ref{fig:case} presents a \textbf{GAIA tabular reasoning task}, where the agent must identify the country with the highest \emph{average medals per athlete} from a CSV file. 
Following Stage I, FluxMem initializes the working context $S_0(q)$ by activating a local subgraph containing semantic knowledge (e.g., CSV parsing and spreadsheet APIs), a relevant episodic trajectory for tabular ranking, and a coarse procedural skill for generic table QA. 
The first execution step succeeds, correctly parsing the file schema from environmental observation $\text{Obs}_1$. 

However, at the next step, the agent incorrectly invokes a spreadsheet visualization API for aggregation, which triggers an \emph{environmental failure} due to rendering limitations. 
FluxMem attributes this failure to a connectivity mismatch and performs targeted topological edits: it \textbf{prunes} the ineffective semantic connection while \textbf{expanding} connectivity toward a more suitable Python data-analysis API, yielding a refined context $S'_1(q)$. 
Subsequently, although aggregation succeeds, self-verification reveals that the inherited procedural skill is overly coarse-grained: it supports ranking existing statistics but fails to compose the required metric (\emph{medals per athlete}). 
This triggers \textbf{Node Reshape}, replacing the skill with a finer-grained statistical aggregation procedure.

%% file: table/main_results_locomo.tex
\begin{table}[t]
\centering
\small

\begin{tabularx}{\columnwidth}{
l|
>{\centering\arraybackslash}X
>{\centering\arraybackslash}X
>{\centering\arraybackslash}X
>{\centering\arraybackslash}X
|>{\centering\arraybackslash}X
}
\toprule
Method & Single Hop & Multi Hop & Temp & Open Domain & AVG \\
\midrule

\multicolumn{6}{c}{\textit{GPT-4.1-mini}} \\
\midrule
Full Context & 87.99 & 80.50 & 71.03 & 58.33 & 81.23 \\
Zep          & 66.90 & 53.70 & 60.20 & 43.80 & 61.60 \\
Mem0         & 71.40 & 68.20 & 56.90 & 47.90 & 66.30 \\
A-Mem        & 77.41 & 61.35 & 71.03 & 50.00 & 71.43 \\
MemoryOS     & 78.95 & 66.67 & 55.45 & 60.42 & 70.65 \\
Nemori       & 87.16 & 77.30 & 76.32 & 55.21 & 81.10 \\
LightMem     & 87.87 & 76.95 & 80.37 & 57.29 & 82.40 \\
MIRIX        & 85.11 & 83.70 & 88.39 & 65.62 & 85.38 \\
EverMemOS    & \textbf{96.67} & \underline{91.84} & \underline{89.72} & \underline{76.04} & \underline{93.05} \\
\midrule
FluxMem   & \underline{95.95} & \textbf{93.26} & \textbf{95.64} & \textbf{90.62} & \textbf{95.06} \\
\midrule

\multicolumn{6}{c}{\textit{Qwen3-30B-A3B-2507-Instruct}} \\
\midrule
Full Context & \underline{87.40} & \underline{69.86} & \underline{51.71} & \underline{57.29} & \underline{74.87} \\
A-Mem        & 67.90 & 57.45 & 27.73 & 43.75 & 56.10 \\
MemoryOS     & 79.51 & 63.12 & 32.09 & 48.96 & 59.81 \\
LightMem     & 82.76 & 68.09 & 50.16 & 52.08 & 71.36 \\
\midrule
FluxMem   & \textbf{96.19} & \textbf{91.49} & \textbf{89.72} & \textbf{87.50} & \textbf{93.44} \\
\bottomrule

\end{tabularx}

\caption{Evaluation results on LoCoMo benchmark. \textbf{Bold} and \underline{underlined} entries indicate the best and second-best results within each group, respectively.}
\label{tab:locomo_results}
\end{table}

%% file: table/main_results_mind2web.tex
\begin{table*}[!t]
  \centering
  \setlength{\tabcolsep}{4pt}
  \small

  \begin{tabular}{lc|cccc|cccc|cccc|cccc}
    \toprule
    \multicolumn{2}{c|}{\multirow{2}{*}{Method}} &
    \multicolumn{4}{c|}{Cross-Task} &
    \multicolumn{4}{c|}{Cross-Website} &
    \multicolumn{4}{c|}{Cross-Domain} &
    \multicolumn{4}{c}{AVG} \\
    \cmidrule(lr){3-6} \cmidrule(lr){7-10} \cmidrule(lr){11-14} \cmidrule(lr){15-18}
    & & EA & AF1 & SSR & SR & EA & AF1 & SSR & SR & EA & AF1 & SSR & SR & EA & AF1 & SSR & SR \\
    \midrule

    \multicolumn{18}{c}{\textit{GPT-4.1-mini}} \\
    \midrule
    \multicolumn{2}{c|}{No Memory$^\ddagger$} &
    34.5 & 64.7 & 28.0 & 2.8 &
    33.0 & 60.4 & 24.6 & 1.1 &
    30.1 & 55.9 & 24.6 & 0.7 &
    31.3 & 58.1 & 25.2 & 1.1 \\

    \multicolumn{2}{c|}{AWM$^\ddagger$} &
    37.4 & 71.4 & 31.6 & 3.6 &
    34.5 & 62.8 & 26.1 & 1.1 &
    30.3 & 56.9 & 24.9 & 1.0 &
    32.2 & 60.4 & 26.3 & 1.5\\
    \multicolumn{2}{c|}{AWM$^\dagger$} &
    % TODO: fill in realistic-setting results
    \underline{64.6} & \underline{82.3} & 56.3 & \underline{9.1} &
    \underline{62.9} & \underline{79.7} & 52.7 & 4.5 &
    50.5 & 69.3 & 43.9 & 2.0 &
    54.8 & \underline{73.1} & 47.4 & 3.7 \\

    \midrule

    \multicolumn{2}{c|}{FluxMem$^\ddagger$} &
    60.7 & 77.5 & \underline{56.7} & 8.1 &
    60.5 & 75.6 & \underline{55.4} & \underline{7.9} &
    \underline{56.2} & \underline{69.4} & \underline{52.1} & \underline{4.3} &
    \underline{57.6} & 71.7 & \underline{53.4} & \underline{5.5} \\

    \multicolumn{2}{c|}{FluxMem$^\dagger$} &
    \textbf{70.4} & \textbf{84.2} & \textbf{60.1} & \textbf{13.5} &
    \textbf{70.2} & \textbf{82.1} & \textbf{58.7} & \textbf{9.2} &
    \textbf{65.2} & \textbf{75.3} & \textbf{55.3} & \textbf{7.2} &

    \textbf{66.8} & \textbf{77.9} & \textbf{56.6} & \textbf{8.6} \\
    \midrule

    \multicolumn{18}{c}{\textit{Gemini-2.5-flash}} \\
    \midrule
    \multicolumn{2}{c|}{No Memory$^\ddagger$} &
    40.7 & 69.5 & 35.6 & 2.8 &
    35.7 & 64.7 & 28.3 & 1.7 &
    22.1 & 37.2 & 18.6 & 1.8 &
    20.5 & 46.9 & 23.1 & 2.0 \\

    \multicolumn{2}{c|}{AWM$^\ddagger$} &
    53.1 & 76.6 & 48.0& 5.6 &
    48.9 & 71.5 & 44.3 & 3.4 &
    41.6 & 60.2 & 38.1 & 1.5 &
    44.7 & 64.8 & 40.8 & 2.5\\
    \multicolumn{2}{c|}{AWM$^\dagger$} &
    % TODO: fill in realistic-setting results
    \underline{67.3} & \underline{82.2} & \underline{60.9} & 8.7 &
    \underline{63.0} & \textbf{77.0} & 55.3 & 4.5 &
    52.0 & 67.0 & 47.9 & 3.4 &
    56.3 & 71.2 & 51.3 & 4.5 \\

    \multicolumn{2}{c|}{ReasoningBank$^\dagger$} &
    52.1 & 60.4 & 44.9 & 4.8 &
    44.3 & 52.6 & 33.9 & 3.3 &
    40.6 & 41.3 & 36.6 & 1.6 &
    43.2 & 46.4 & 37.8 & 2.4 \\

    \midrule
    
    \multicolumn{2}{c|}{FluxMem$^\ddagger$} &
    61.6 & 77.4 & 57.0 & \underline{9.6} &
    59.4 & \underline{72.8} & \underline{56.1} & \underline{7.5} &
    \underline{58.5} & \underline{70.1} & \textbf{55.9} & \underline{5.0} &
    \underline{59.2} & \underline{71.8} & \underline{56.1} & \textbf{6.2} \\

    \multicolumn{2}{c|}{FluxMem$^\dagger$} &
    \textbf{71.3} & \textbf{83.0} & \textbf{61.3} & \textbf{11.1} &
    \textbf{65.9} & \textbf{77.0} & \textbf{59.7} & \textbf{8.5} &
    \textbf{64.8} & \textbf{70.3} & \underline{54.6} & \textbf{7.0} &
    \textbf{66.2} & \textbf{73.6} & \textbf{56.5} & \underline{5.9} \\

    \bottomrule
  \end{tabular}%

  \caption{Evaluation results on Mind2Web. $\dagger$ denotes the filtered setting with manual removal of non-golden HTML elements; $\ddagger$ denotes the realistic setting without any pre-filtering (see Appendix~\ref{appendix:implementation_details} for details). \textbf{Bold} and \underline{underline} denote the best and second-best results.}

  \label{tab:mind2web_results}
  \vspace{-10pt}
\end{table*}

%% file: table/main_results_gaia.tex
\begin{table*}[!t]
\centering
\footnotesize

\setlength{\tabcolsep}{3pt}
\resizebox{\textwidth}{!}{
\begin{tabular}{l@{\hspace{4pt}}c|cccc}
\toprule
\midrule
\textbf{Framework} & \textbf{Model} & \textbf{Avg.} & \textbf{Level 1} & \textbf{Level 2} & \textbf{Level 3} \\
\midrule
\multicolumn{6}{l}{\textit{Closed-source Agent Frameworks}} \\
\midrule
Langfun & Claude 3.7 etc. & 71.52 & \underline{83.02} & 68.60 & \textbf{57.69} \\
OpenAI Deep Research & OpenAI o1, o3 etc. & 67.36 & 74.29 & 69.06 & 47.60 \\
\midrule
\multicolumn{6}{l}{\textit{Open-Source Agent Frameworks}} \\
\midrule
Magnetic-1 & OpenAI o1 etc. & 46.06 & 56.60 & 46.51 & 23.08 \\
Agent KB & GPT-4.1 & 61.21 & 79.25 & 58.14 & 34.62 \\
Alita & Claude-3.7-Sonnet,GPT-4o & 72.73 & 81.13 & \textbf{75.58} & 46.15 \\
\midrule
Smolagents & GPT-4.1 & 55.15 & 67.92 & 53.49 & 34.62 \\
Smolagents & GPT-5-mini & 55.75 & 69.81 & 54.65 & 30.77 \\
\midrule
Flash-Searcher & GPT-5-mini & 69.09 & 79.25 & 69.77 & 46.15 \\
Flash-Searcher & Kimi K2 & 52.12 & 58.49 & 52.33 & 34.62 \\
Flash-Searcher & DeepSeek V3.2 & 60.61 & 79.25 & 53.49 & 46.15 \\
\midrule
MemEvolve & GPT-5-mini & $69.09 \rightarrow \underline{73.33}$ \textcolor[rgb]{0.35,0.65,0.45}{$\uparrow$4.24} & $79.25 \rightarrow$ \underline{83.02} \textcolor[rgb]{0.5,0.75,0.6}{$\uparrow$3.77} & $69.77 \rightarrow \underline{73.26}$ \textcolor[rgb]{0.5,0.75,0.6}{$\uparrow$3.49} & $46.15 \rightarrow \underline{53.85}$ \textcolor[rgb]{0.2,0.55,0.3}{$\uparrow$7.70} \\
MemEvolve & Kimi K2 & $52.12 \rightarrow 61.21$ \textcolor[rgb]{0.1,0.45,0.2}{$\uparrow$9.09} & $58.49 \rightarrow 67.92$ \textcolor[rgb]{0.1,0.45,0.2}{$\uparrow$9.43} & $52.33 \rightarrow 63.95$ \textcolor[rgb]{0.05,0.35,0.15}{$\uparrow$11.62} & $34.62 \rightarrow 38.46$ \textcolor[rgb]{0.5,0.75,0.6}{$\uparrow$3.84} \\
MemEvolve & DeepSeek V3.2 & $60.61 \rightarrow 67.88$ \textcolor[rgb]{0.2,0.55,0.3}{$\uparrow$7.27} & $79.25 \rightarrow$ \underline{83.02} \textcolor[rgb]{0.5,0.75,0.6}{$\uparrow$3.77} & $53.49 \rightarrow 63.95$ \textcolor[rgb]{0.08,0.4,0.18}{$\uparrow$10.46} & $46.15 \rightarrow 50.00$ \textcolor[rgb]{0.5,0.75,0.6}{$\uparrow$3.85} \\
\midrule
FluxMem & GPT-5-mini & $69.09 \rightarrow$ \textbf{76.36} \textcolor[rgb]{0.2,0.55,0.3}{$\uparrow$7.27} & $79.25 \rightarrow$ \textbf{88.68} \textcolor[rgb]{0.1,0.45,0.2}{$\uparrow$9.43} & $69.77 \rightarrow$ \textbf{75.58} \textcolor[rgb]{0.3,0.6,0.4}{$\uparrow$5.81} & $46.15 \rightarrow \underline{53.85}$ \textcolor[rgb]{0.2,0.55,0.3}{$\uparrow$7.70} \\
FluxMem & Kimi K2 & $52.12 \rightarrow 64.85$ \textcolor[rgb]{0.05,0.35,0.15}{$\uparrow$12.73} & $58.49 \rightarrow 77.36$ \textcolor[rgb]{0.0,0.25,0.08}{$\uparrow$18.87} & $52.33 \rightarrow 62.79$ \textcolor[rgb]{0.08,0.4,0.18}{$\uparrow$10.46} & $34.62 \rightarrow 46.15$ \textcolor[rgb]{0.05,0.35,0.15}{$\uparrow$11.53} \\
FluxMem & DeepSeek V3.2 & $60.61 \rightarrow 70.30$ \textcolor[rgb]{0.1,0.45,0.2}{$\uparrow$9.69} & $79.25 \rightarrow 81.13$ \textcolor[rgb]{0.55,0.8,0.65}{$\uparrow$1.88} & $53.49 \rightarrow 69.77$ \textcolor[rgb]{0.0,0.25,0.08}{$\uparrow$16.28} & $46.15 \rightarrow 50.00$ \textcolor[rgb]{0.5,0.75,0.6}{$\uparrow$3.85} \\
\midrule
\bottomrule
\end{tabular}}
\caption{Evaluation results on GAIA benchmark. Color-coded arrows denote improvements relative to the Flash-Searcher baseline, with deeper green shades indicating larger score improvements. \textbf{Bold} and \underline{underline} indicate the best and second-best final results across all methods, respectively.}
\label{tab:gaia_results}

\end{table*}

%% file: section/relatedwork.tex
\section{Related Work}

\paragraph{Hierarchical Structured Memory Systems.}

Hierarchical memory systems organize memory units through specific topologies. 
Tree-based memories abstract data into hierarchical levels~\citep{a2024enhancinglongtermmemoryusing, ye2025h2rhierarchicalhindsightreflection}, while graph-based structures offer greater connectivity dynamics~\citep{long2025seeinglisteningrememberingreasoning}. 
Pyramid mechanisms construct multi-level abstractions to facilitate coarse-to-fine querying~\citep{han2025retrievalaugmentedgenerationgraphsgraphrag, rasmussen2025zeptemporalknowledgegraph}. 
Alternatively, heterogeneous multi-layer structures partition memory into distinct modules~\citep{zhang2025memevolvemetaevolutionagentmemory,structmem} or levels specialized for specific information types or functions~\citep{zhang2025gmemorytracinghierarchicalmemory, gutiérrez2025hipporagneurobiologicallyinspiredlongterm, gutiérrez2025ragmemorynonparametriccontinual}.

\paragraph{Self Evolving Agent Memory.}

Agent self-evolution is typically driven by memory evolving~\citep{gao2026surveyselfevolvingagentswhat, fang2025comprehensivesurveyselfevolvingai, shinn2023reflexion}, which can be categorized into three paradigms. 
\textbf{(i)} Works such as Expel~\citep{zhao2024expel}, AWM~\citep{wang2024agent}, and ReasoningBank~\citep{ouyang2025reasoningbankscalingagentselfevolving} maintain a contextual memory repository~\citep{liu2025contextualexperiencereplayselfimprovement, wei2025evomemorybenchmarkingllmagent, cai2025flexcontinuousagentevolution}, distilling experiences from historical trajectories to enhance agent capabilities~\citep{qiu2025alitageneralistagentenabling}. 
These methods vary in the granularity of trajectory processing~\citep{fang2026mempexploringagentprocedural}. 
In addition to successful experiences, some approaches~\citep{tang2025agentkbleveragingcrossdomain, cao2025remembermerefineme} incorporate failure cases, while others~\citep{wu2025evolverselfevolvingllmagents} introduce iterative evolution mechanisms. 
\textbf{(ii)} Other works focus on parametric memory. 
SEAL~\citep{zweiger2025selfadaptinglanguagemodels} explores the potential of agent self-training, while AgentEvolver~\citep{zhai2025agentevolverefficientselfevolvingagent} and Agent0~\citep{xia2025agent0unleashingselfevolvingagents} concentrate on data acquisition and reward design. 
The evolutionary mechanisms primarily include SFT~\citep{zhou2025mementofinetuningllmagents}, RL~\citep{zhang2026memrlselfevolvingagentsruntime} and other paradigms such as early experience~\citep{shi2025youtuagentscalingagentproductivity}.
\textbf{(iii)} Some works modify model architectures to enable deeper memory-augmented evolving. 
This is achieved by either introducing additional parameters as external memory~\citep{wang2024memoryllmselfupdatablelargelanguage, wang2024wise} to manage information acquisition and forgetting, or proposing novel architectures~\citep{behrouz2024titanslearningmemorizetest, behrouz2025itsconnectedjourneytesttime, behrouz2025nestedlearningillusiondeep} to enhance the inherent memory capacity of base models. 

%% file: section/conclusion.tex
\section{Conclusion}
We introduced \textbf{FluxMem}, an evolutionary framework conceptualizing agent memory as dynamic connectivity. By a three-phase evolution, \textbf{FluxMem} enables autonomous memory adaptation. SOTA results across LoCoMo, Mind2Web and GAIA, provides a principled foundation for self-evolving agents in dynamic environments. 

%% file: section/limitations.tex
\section*{Limitations}
Despite the demonstrated effectiveness, several limitations in our experimental design warrant acknowledgment:
\textbf{Computational Overhead of Closed-Loop Operations.} Stages II and III rely on iterative LLM calls for context verification, topological editing, and skill induction. Our current evaluation prioritizes task success and convergence, but does not systematically measure the associated latency, API cost, or token consumption, which are critical for real-time or resource-constrained deployments.
\textbf{Static Benchmark Protocols.} Experiments are conducted on pre-collected, static datasets (LoCoMo, Mind2Web, GAIA). While diverse, these benchmarks do not fully simulate continuous, open-world distribution shifts or streaming environments where task boundaries blur and memory decay must be actively managed alongside evolution.
\textbf{Hyperparameter Sensitivity.} The framework introduces several control thresholds (e.g., refinement rounds $T$, PEMS convergence threshold $\epsilon$, retrieval top-$k$). Our ablations focus on component efficacy but lack a comprehensive sensitivity analysis across different model backbones and highly heterogeneous domains. Future work should systematically evaluate the robustness of these parameters under varying computational budgets.
\textbf{Offline Consolidation Scheduling.} Stage III is executed offline in periodic batches. The current experimental setup does not evaluate dynamic scheduling strategies or the trade-off between consolidation frequency and online performance degradation, which are essential for practical lifelong agent deployment.

%% file: section/Use_of_AI.tex
\section*{Use of AI Assistants}
During the preparation of this manuscript, AI language models were utilized exclusively for linguistic refinement, grammar correction, and stylistic polishing. The core research concepts, experimental design, methodological implementation, narrative structure, and all figures/illustrations were independently conceived, developed, and verified by the human authors. The authors retain full responsibility for the scientific accuracy, originality, and integrity of the work.